\documentclass[11pt]{article}

\usepackage[preprint]{acl}

\usepackage{times}
\usepackage{latexsym}

\usepackage[T1]{fontenc}

\usepackage[utf8]{inputenc}

\usepackage{microtype}

\usepackage{inconsolata}

\usepackage{graphicx}
\usepackage{tikz}
\usepackage{amsmath}
\usepackage{booktabs}
\usepackage{colortbl}
\usepackage{multirow} 
\usepackage{subfigure}
\usepackage{subcaption}

\definecolor{lightgreen}{RGB}{220,255,220}
\definecolor{lightred}{RGB}{255,220,220}
\definecolor{lightblue}{RGB}{220,240,255}
\definecolor{lightorange}{RGB}{255,235,200}

\usepackage{tcolorbox}
\tcbuselibrary{listings, skins, breakable}

\newtcolorbox[auto counter, number within=section]{examplebox}[2][]{%
  colback=blue!5!white,
  colframe=blue!50!black,
  title=#2
}
\newtcolorbox{examplegroup}[2][]{%
  colback=gray!5!white,
  colframe=gray!50!black,
  fonttitle=\bfseries,
  breakable,
  #1
}
\tcbset{
  promptbox/.style={
    colback=blue!2!white,   
    colframe=blue!30!black,
    fonttitle=\bfseries,
    breakable,
    boxrule=0.5pt,
    left=3mm,
    right=3mm,
    top=2mm,
    bottom=2mm,
    before skip=2em,   
    after skip=2em     
  }
}

\title{LiveCLKTBench: Towards Reliable Evaluation of Cross-Lingual Knowledge Transfer in Multilingual LLMs}

\author{
  \textbf{Pei-Fu Guo\textsuperscript{1}},
  \textbf{Yun-Da Tsai\textsuperscript{1}},
  \textbf{Chun-Chia Hsu\textsuperscript{1}},
  \textbf{Kai-Xin Chen\textsuperscript{1}},
  \textbf{Ya-An Tsai\textsuperscript{1}},
\\
  \textbf{Kai-Wei Chang\textsuperscript{2}},
  \textbf{Nanyun Peng\textsuperscript{2}},
  \textbf{Mi-Yen Yeh\textsuperscript{3}},
  \textbf{Shou-De Lin\textsuperscript{1,4}}
\\
\\
  \textsuperscript{1}National Taiwan University
  \textsuperscript{2}University of California, Los Angeles
\\
  \textsuperscript{3}Academia Sinica, Taiwan
  \textsuperscript{4}NTU AI-CoRE
\\
  \small{
    \textbf{Correspondence:} \href{mailto:r12922217@csie.ntu.edu.tw}{r12922217@csie.ntu.edu.tw}
  }
}

\begin{document}

\maketitle
\begin{abstract} 
Evaluating cross-lingual knowledge transfer in large language models (LLMs) is challenging, as correct answers in a target language may arise either from genuine transfer or from prior exposure during pre-training.
We present LiveCLKTBench, an automated generation pipeline specifically designed to isolate and measure cross-lingual knowledge transfer. Our pipeline identifies self-contained, time-sensitive knowledge entities from real-world domains, filters them based on temporal occurrence, and verifies them against the model’s knowledge. The documents of these valid entities are then used to generate factual questions, which are translated into multiple languages to evaluate transferability across linguistic boundaries.
Using LiveCLKTBench, we evaluate several LLMs across five languages and observe that cross-lingual transfer is strongly influenced by linguistic distance and often asymmetric across language directions. While larger models improve transfer, the gains diminish with scale and vary across domains.
These findings provide new insights into multilingual transfer and demonstrate the value of LiveCLKTBench as a reliable benchmark for future research.\footnote{Code and data are available at \href{https://github.com/0Frett/Live-CLKT-BENCH}{link}.}
\end{abstract}

\section{Introduction}
\label{sec:intro}
\begin{figure*}[t]
    \centering
    \includegraphics[width=1.0\linewidth]{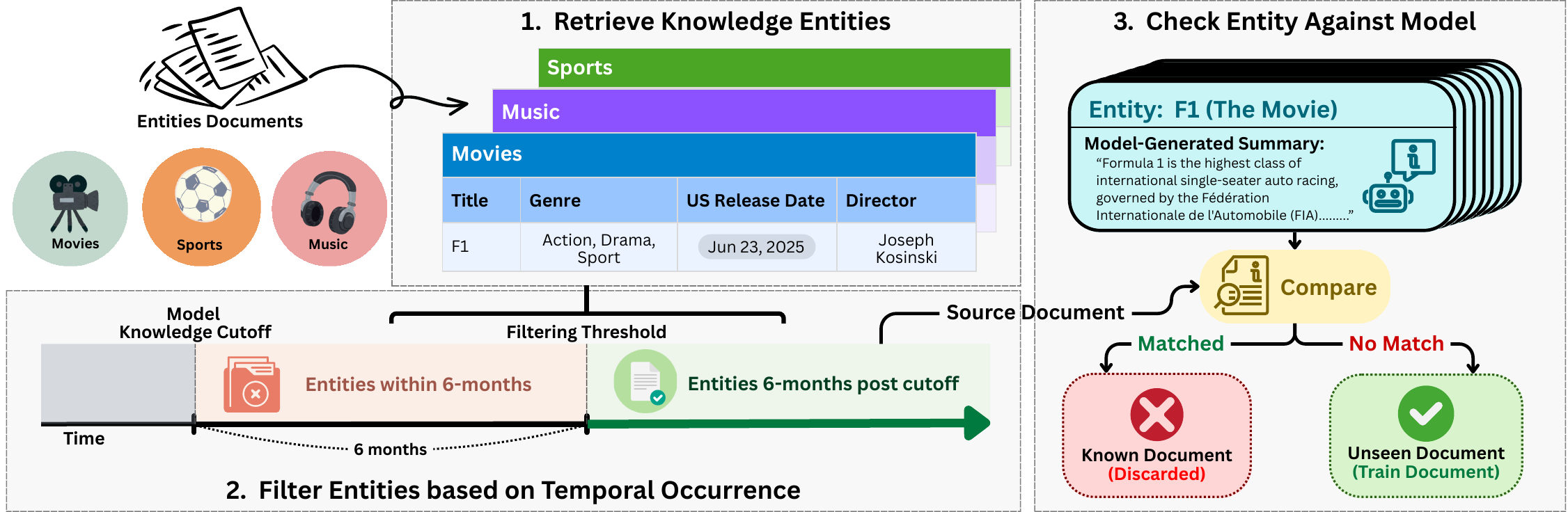}
    \caption{
        \textbf{Leakage Prevention in LiveCLKTBench.} 
            The pipeline prevents data leakage by selecting valid knowledge entities that contain facts unknown to pretrained models.
            Specifically, it identifies independent, time-sensitive real-world entities, filters them by temporal occurrence, and cross-checks them against model outputs to eliminate any entities already known to pretrained models.
    }
    \label{fig:leakage}
\end{figure*}

As large language models (LLMs) continue to grow in scale and capability, a central question arises: \emph{How can they serve users globally and equitably?}
Ideally, an LLM should be able to transfer knowledge acquired in one language to others, rather than relearning the same facts separately across languages — a property known as \textit{cross-lingual knowledge transfer}~\citep{pmlr-v119-hu20b, lauscher-etal-2020-zero, hedderich-etal-2020-transfer}. 

However, reliably evaluating this ability remains challenging.
As LLMs are pretrained on massive multilingual corpora that may already contain the same factual knowledge in multiple languages, it is often unclear whether correct answers in the target language reflect genuine cross-lingual transfer or merely memorization of previously seen information. 
To address this issue, we identify three essential properties that a reliable cross-lingual transfer benchmark should satisfy.

\textbf{(1) Leakage-Free Evaluation.}
A major obstacle in evaluating cross-lingual knowledge transfer is contamination and data leakage~\citep{ahuja2024contaminationreportmultilingualbenchmarks}.
When pretrained corpora already include the same knowledge in multiple languages, models can appear to “transfer” information they have merely memorized.
A leakage-free benchmark must therefore ensure that the tested knowledge is unseen in the target language, so that correct answers truly reflect transfer.

\textbf{(2) Grounding in Real-World Knowledge.} Some benchmarks employ synthetic or fictitious data to control knowledge injection and reduce contamination~\citep{maini2024tofutaskfictitiousunlearning, maheshwari2024efficacysyntheticdatabenchmark}. While this strategy offers strong controllability, recent works have shown that fabricated facts can conflict with pre-existing knowledge and degrade  performance~\citep{wu2024akewassessingknowledgeediting, jan2025datadopingtrueintelligence, chen-etal-2024-unveiling-flaws}. Consequently, when a model fails to answer a fictitious question in a particular language, the failure may not stem from a lack of transferability, but rather from the artificial knowledge being too weakly or even adversarially connected to existing knowledge to support meaningful transfer. We therefore argue that benchmarks grounded in real documents better reflect practical scenarios, where models must absorb genuine knowledge and transfer it across languages while leveraging their existing background knowledge.

\textbf{(3) Frequent Knowledge Update.} As LLMs evolve rapidly, static benchmarks become outdated: newer models may already encode most benchmark facts, causing score saturation and reducing diagnostic value~\footnote{\url{https://r0bk.github.io/killedbyllm/}}. Recent initiatives such as \textsc{RealTimeQA}~\citep{kasai2024realtimeqawhatsanswer} and \textsc{LiveBench}~\citep{white2025livebenchchallengingcontaminationlimitedllm} highlight the importance of designing benchmarks that are regularly refreshed with new data to prevent staleness.

Building on these principles, we introduce \emph{LiveCLKTBench}, an automated pipeline for generating realistic, contamination-free, and continuously refreshable benchmarks for cross-lingual knowledge transfer.
As shown in Figure~\ref{fig:leakage}, given a target model, our pipeline minimizes the risk of data leakage through the following strategies:  
\begin{enumerate}
    \item Identifying independent, time-sensitive \emph{knowledge entities} from three frequently updated real-world domains—\emph{movies}, \emph{music} and \emph{sports} (e.g., a new release movie, a recent baseball game match).
    \item Filtering entities based on temporal occurrence, retaining only those appearing at least six months after the model’s knowledge cutoff to avoid potential prior exposure.  
    \item Verifying each entity by prompting the model for a factual summary; entities whose responses match the real-world \emph{source document} are treated as known and discarded.
\end{enumerate}
Together, these steps ensure that the resulting QA pairs are contamination-free, grounded in real-world knowledge, and provide a reliable foundation for evaluating cross-lingual knowledge transfer.

For the retained knowledge entities, the pipeline generates factual multiple-choice questions that are (i) explicitly grounded in the corresponding source documents, and (ii) whose correct answers become available only once the event has taken place (e.g., the final score of a sports match). The verified questions and their corresponding source documents are translated into the evaluation languages. During evaluation, models are post-trained on source-language documents and tested on QA pairs in other languages. Because the knowledge is unseen by the model, the benchmark provides a direct measure of knowledge transfer, eliminating concerns about memorization from pretraining.

In addition to ensuring data integrity, LiveCLKTBench offers full automation and configurability, accommodating new model releases and flexible evaluation parameters.
By simply specifying the target models and evaluation languages, researchers can conduct frequent and reproducible evaluations with minimal manual intervention.

Using LiveCLKTBench, we evaluate a series of open-source multilingual LLMs across five languages.
Our analysis shows that current models still have substantial room for improvement: transferability is strongly influenced by linguistic distance and often asymmetric across language directions.
While larger models generally exhibit better transfer, their gains diminish with scale and vary across domains.

Together, these findings provide new insights into cross-lingual knowledge transfer and highlight the value of LiveCLKTBench as a reliable and sustainable testbed for studying and improving multilingual LLMs.

\section{Related Work}
\label{sec:related_work}

\subsection{Multilingual Benchmarks}
A variety of multilingual benchmarks have been proposed to evaluate model performance across many languages~\citep{wu2025bitterlessonlearned2000}. 
Some focus on general natural-language tasks such as classification, question answering, or summarization (e.g., XTREME~\citep{hu2020xtreme}, XGLUE~\citep{liang2020xglue}, MEGA~\citep{ahuja2023megamultilingualevaluationgenerative}), 
while others emphasize factual knowledge access in multiple languages, such as M3Exam~\citep{zhang2023m3exammultilingualmultimodalmultilevel}, AGI-Eval~\citep{zhong2023agievalhumancentricbenchmarkevaluating}, and Global-MMLU~\citep{singh2025globalmmluunderstandingaddressing}. 
These benchmarks provide broad coverage of multilingual capabilities but do not directly isolate cross-lingual transfer.  

\subsection{Cross-lingual Transfer Benchmarks}
\emph{Cross-lingual transfer} benchmarks test whether \emph{knowledge} or \emph{skills} learned in one language generalize to others. 
Here, \emph{skill transfer} refers to the generalization of abilities such as summarization or instruction following to unseen languages~\citep{shaham-etal-2024-multilingual, chai2024xcotcrosslingualinstructiontuning, asai2023buffetbenchmarkinglargelanguage}, 
whereas \emph{knowledge transfer} evaluates the reproduction of factual information acquired in one language when queried in another.

Recent studies analyze the unique challenges of \emph{cross-lingual knowledge transfer}, such as language transfer asymmetry and knowledge representation barriers~\citep{rajaee2024analyzingevaluationcrosslingualknowledge, chua2025crosslingualcapabilitiesknowledgebarriers, litschko2025crossdialectinformationretrievalinformation,yao2024datacontaminationcrosslanguage}.
Complementing these analyses, several dedicated benchmarks have been proposed.
For example, ECLeKTic~\citep{goldman2025eclektic} benchmarks cross-lingual knowledge transfer by using Wikipedia articles common in one language but rare in others, ensuring answers reflect genuine transfer.
Other benchmarks focus on multilingual knowledge editing, evaluating whether updates introduced in one language propagate to others~\citep{nie2025bmike53investigatingcrosslingualknowledge, wei2025mlakemultilingualknowledgeediting}.

Another important dimension is the \emph{evaluation protocol}. 
Some benchmarks use zero-shot evaluation, querying models directly in target languages~\citep{malkin-etal-2022-balanced, goldman2025eclektic}, 
while others adopt a fine-tune–then-test approach, where models are trained in one language and tested in others~\citep{shaham-etal-2024-multilingual}. 
Work also distinguishes between \emph{parametric knowledge}, stored in model parameters~\citep{goldman2025eclektic, rajaee2024analyzingevaluationcrosslingualknowledge}, 
and \emph{contextual knowledge}, provided through in-context examples at inference time~\citep{mondshine2025englishimpactprompttranslation, asai2021xorqacrosslingualopenretrieval}.

\subsection{Relation to Prior Work}
Among prior efforts, ECLeKTic~\citep{goldman2025eclektic} is most closely related to our work, as it evaluates factual transfer across languages while reducing leakage by selecting documents that are more common in one language than another. While this strategy offers partial control, leakage cannot be fully ruled out, and reliance on Wikipedia limits coverage of dynamic knowledge. In contrast, LiveCLKTBench injects new knowledge via post-training on a single source language, so correct answers in other languages must arise from genuine transfer. By sourcing documents from rapidly updating domains and allowing configurable time windows, it produces more diverse and extensible benchmarks that stay relevant as models evolve.

\section{Methodology}
\label{sec:dataset}

\begin{figure*}[t]
    \centering
    \includegraphics[width=1.0\linewidth]{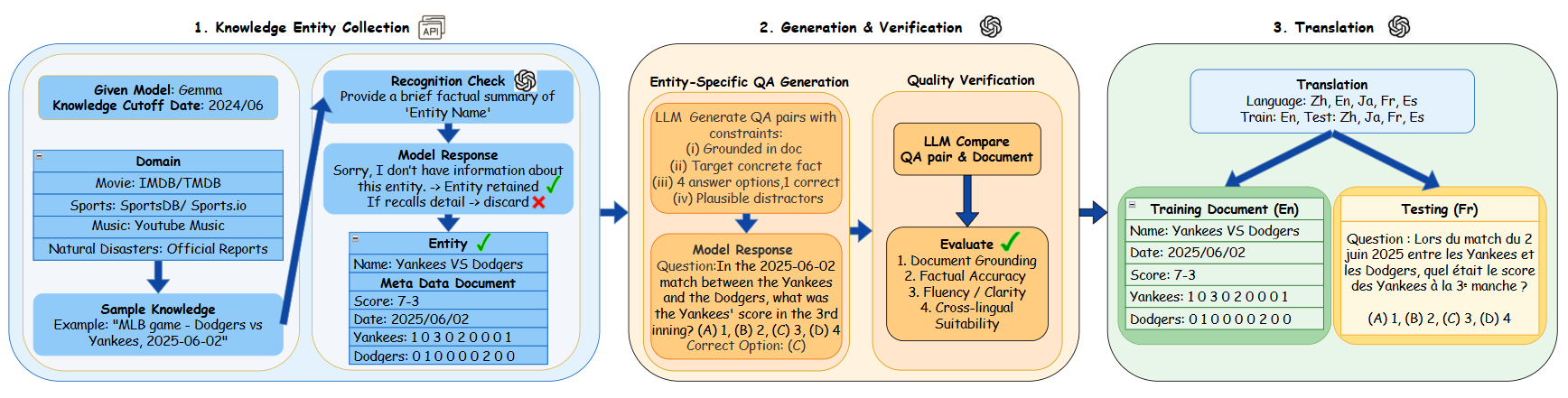}
    \caption{\textbf{LiveCLKTBench Pipeline.}
The generation process consists of four stages:
(1) collecting independent, time-sensitive knowledge entities;
(2) generating document-grounded question–answer pairs;
(3) verifying data quality using a verifier LLM; and
(4) translating verified questions into multiple languages for evaluation.
}
    \label{fig:data_gen}
\end{figure*}

\subsection{Benchmark Generation Pipeline}
\label{sec:construction}
Each LiveCLKTBench benchmark consists of two components:  
(i) a set of source-language training documents that introduce knowledge previously unseen by the model, and  
(ii) a multilingual test set of document-grounded factual questions derived from those documents.  
As illustrated in Figure~\ref{fig:data_gen}, the pipeline proceeds through four main stages, each applying strict constraints to prevent knowledge contamination. Starting with a target model and the evaluation language set, we describe each generation stage in detail below.

\paragraph{Knowledge Entity Collection.}
\label{sec:construction:entity_collect}
The first stage collects \emph{knowledge entities} that serve as the basis for generating test set factual QA pairs. Our pipeline focus on three domains—\emph{movies}, \emph{music}, and \emph{sports}. They are chosen because they frequently produce entities with fresh facts that satisfy two key properties:

\begin{itemize}
    \item \textbf{Independence.} Each entity is self-contained, and its facts cannot be inferred from other events. For example, the outcome of one baseball game does not depend on the results of earlier games, allowing for clean separation between training and test instances.  
    \item \textbf{Time-sensitivity.} The factual outcome itself (e.g., which team won or the exact score) becomes known only after the match has taken place, ensuring that such information could not have been included in pretraining data.  
\end{itemize}

Entities are first retrieved from each domain using specific databases such as IMDB/TMDB (movies), YouTube Music (music) and SportsDB/Sports.io (sports). To avoid contamination, we apply a strict temporal filter, where only entities appearing at least six months after the model’s pretraining cutoff are considered. This reduces the risk of early exposure through trailers, previews, press leaks, or other pre-release publicity. The time window is configurable to match different model knowledge cutoffs and experimental needs.

After sampling, we perform a conservative check against the target model’s \emph{pretrained} checkpoint to ensure each entity is genuinely unseen. The model is prompted with a short, unambiguous request: \texttt{"Provide a brief factual summary of `<ENTITY NAME>'."} The response is then compared to the retrieved source document using an LLM judge. If the response includes concrete facts consistent with the document, the entity is flagged as recognized and discarded.

For all remaining entities, we retrieve a canonical source document (e.g., a baseball match report, IMDB film page, music release metadata) to serve as the factual grounding for downstream QA generation. This multi-step procedure ensures that the retained entities and their corresponding documents constitute genuinely unseen knowledge, establishing a robust and contamination-free foundation for benchmark construction. Retrieved documents examples are provided in Appendix~\ref{app:doc_metadata}.

\paragraph{Entity-Specific QA Generation.}  
\label{sec:construction:qa_gen}

For each retained entity, we generate multiple-choice questions that are factual and grounded in the corresponding source document using an LLM (GPT-4o-mini). The generation prompts enforce strict constraints: (i) every question must be explicitly grounded in the document; (ii) each question must target a concrete fact rather than subjective interpretation; (iii) exactly four candidate answers must be provided, with only one correct option; and (iv) distractors must be plausible but incorrect.

Because these questions focus on facts that emerge only after the event occurs, they are highly unlikely to appear in any pretraining data. These constraints ensure that the generated QAs are verifiable, unambiguous, and sufficiently challenging, providing a reliable foundation for evaluating cross-lingual knowledge transfer. 
For instance, a valid example that satisfies the four criteria is:
\begin{examplebox}{Sports QA}
\texttt{In the sports game 'Los Angeles Dodgers vs Toronto Blue Jays' at 2025-08-09, what was the final score?}  \\
(A.) 5 -- 1 
(B.) 3 -- 2 
(C.) 4 -- 0 
(D.) 6 -- 5  
\end{examplebox}
Additional QA examples are provided in Appendix~\ref{app:qa_example}.


\paragraph{Quality Verification.}
\label{sec:construction:qa_verify}
To ensure that generated questions meet the required standards, a verifier LLM checks each QA pair against the original document and explicitly validates the four criteria above. Any question that violates at least one criterion is discarded. This guarantees that every retained test instance is document-grounded, factually accurate, and suitable for assessing transferability. Additionally, we validate the quality of verifier-approved questions by human inspection, confirming a precision of 95\%.  
Details of verification procedure are provided in Appendix~\ref{app:data_verify}.

\paragraph{Translation.}  
\label{sec:construction:translation}
Finally, the verified QAs and their associated documents are translated into the evaluation languages specified by the user. The resulting benchmark consists of a \textbf{train set}, containing source-language documents for knowledge injection, and a \textbf{test set}, containing multilingual factual QAs that probe cross-lingual transferability. We also conduct human evaluation to assess translation quality, as detailed in Appendix~\ref{app:translation_quality}.

\paragraph{Language Coverage.}  
Since our framework relies on state-of-the-art LLMs (e.g., GPT-4o-mini) for translation, language coverage is bounded by the underlying model’s capabilities. In principle, evaluation data can be generated for any language with sufficient translation quality, enabling broader and more comprehensive cross-lingual evaluation.

Through these stages, LiveCLKTBench produces benchmarks that are leakage-free, configurable, and continuously extensible, ensuring evaluations remain independent of pretraining exposure. Detailed prompts for QA generation, verification, and translation are provided in Appendix~\ref{app:prompts}.

\subsection{Evaluation Protocol}
\label{sec:protocol}
Unlike existing multilingual benchmarks that probe pretrained models in a zero-shot setting, LiveCLKTBench evaluates cross-lingual transferability \emph{after knowledge injection}. The protocol focuses on two crucial steps, ensuring that results reflect true transfer:
(1) Knowledge Injection: Inject new knowledge by post-training the model on source-language documents.  
(2) Transfer Evaluation: Evaluate the post-trained model on test QA pairs in the target languages.  
This setup guarantees that any correct prediction in a target language arises solely from successful cross-lingual knowledge transfer.

LiveCLKTBench supports two primary scenarios that allow researchers to analyze transferability from different perspectives:
\begin{enumerate}
    \item \textbf{Model Comparison:} Compare multiple models using the same post-training strategy to determine which exhibits stronger cross-lingual transfer capabilities.
    \item \textbf{Strategy Comparison:} For a single model, test different post-training methods (e.g., continual pretraining~\citep{ke2023continualpretraininglanguagemodels}, supervised fine-tuning~\citep{mecklenburg2024injectingnewknowledgelarge}, knowledge editing~\citep{wu2024akewassessingknowledgeediting}) to identify approaches that yield better transferability.
\end{enumerate}
By supporting these scenarios, LiveCLKTBench enables both model-level and method-level analyses of cross-lingual knowledge transfer.

\subsection{Evaluation Metrics}
\label{sec:eval_metric}

To quantify cross-lingual transferability, we follow the metric definitions introduced by \citet{goldman2025eclektic}. Let $\mathcal{L}$ denote the set of evaluation languages. For each ordered pair $(L_{\text{train}}, L_{\text{test}})$ with $L_{\text{train}} \neq L_{\text{test}}$, model predictions can be categorized using the contingency matrix in Table~\ref{tab:confusion}.  

\begin{table}[h]
    \centering

    \begin{tabular}{@{}l|c|c@{}}
        \toprule
         & $L_{\text{test}}$ Correct & $L_{\text{test}}$ Wrong \\
        \midrule
        $L_{\text{train}}$ Correct & \cellcolor{lightgreen}$A$ & \cellcolor{lightorange}$B$ \\
        $L_{\text{train}}$ Wrong   & \cellcolor{lightblue}$C$ & \cellcolor{lightred}$D$ \\
        \bottomrule
    \end{tabular}
    \caption{Contingency matrix for $(L_{\text{train}}, L_{\text{test}})$. 
        $A$: successful transfer, 
        $B$: failed transfer, 
        $C$: cross-lingual inconsistency, 
        $D$: complete failure.}
    \label{tab:confusion}
\end{table}

This matrix distinguishes four possible outcomes: 
$A$ — correct in both source and target languages (\textit{successful transfer}); 
$B$ — correct in source but wrong in target (\textit{failed transfer}); 
$C$ — wrong in source but correct in target (\textit{cross-lingual inconsistency}); and 
$D$ — wrong in both (\textit{fail to learn}). From these outcomes we derive two complementary metrics.

\paragraph{Overall Success.}  
This metric measures how often the model answers questions correctly in both the source and target languages. It reflects the combined ability to acquire knowledge during source-language post-training and to consistently reproduce it across languages:  
\begin{equation}
    \text{Overall}(L_{\text{train}},L_{\text{test}}) = \frac{A}{A+B+C+D}.
\end{equation}

\paragraph{Transfer Success.}  
This metric conditions on cases where the source language answer is correct and evaluates the proportion that are also answered correctly in the target language. It directly quantifies the reliability of transferring acquired knowledge:  
\begin{equation}
    \text{Transfer}(L_{\text{train}},L_{\text{test}}) = \frac{A}{A+B}.
\end{equation}

\medskip
Together, these metrics separate two dimensions of cross-lingual performance: 
\textbf{Overall Success} captures how frequently knowledge is jointly expressed across languages, while \textbf{Transfer Success} isolates the likelihood of successful transfer once the source knowledge has been learned.


\begin{table*}[t]
\centering
\begin{tabular}{@{}llccc@{}}
    \toprule
    \textbf{Model} & \textbf{Domain} & \shortstack{\textbf{Overall} \\  \textbf{(Score $\pm$ Std)}} & \shortstack{\textbf{Transfer} \\  \textbf{(Score $\pm$ Std)}} & \shortstack{\textbf{Average over Domain} \\ \textbf{(Overall / Transfer)}} \\
    \midrule
    \multirow{3}{*}{Gemma-2-9b} 
    & music & $0.494 \pm 0.110$ & $0.807 \pm 0.112$ & \multirow{3}{*}{\textbf{0.414} / 0.735} \\
    & movie & $0.483 \pm 0.098$ & $0.778 \pm 0.142$ & \\
    & sports & $0.265 \pm 0.111$ & $0.620 \pm 0.231$ & \\
    \cmidrule{1-5}
    \multirow{3}{*}{Qwen2.5-7B} 
    & music & $0.467 \pm 0.103$ & $0.808 \pm 0.123$ & \multirow{3}{*}{0.387 / \textbf{0.747}} \\
    & movie & $0.452 \pm 0.083$ & $0.794 \pm 0.132$ & \\
    & sports & $0.243 \pm 0.097$ & $0.637 \pm 0.216$ & \\
    \cmidrule{1-5}
    \multirow{3}{*}{Ministral-8B} 
    & music & $0.357 \pm 0.107$ & $0.744 \pm 0.149$ & \multirow{3}{*}{0.304 / 0.669} \\
    & movie & $0.360 \pm 0.097$ & $0.711 \pm 0.194$ & \\
    & sports & $0.195 \pm 0.102$ & $0.551 \pm 0.258$ & \\
    \cmidrule{1-5}
    \multirow{3}{*}{Mistral-Nemo} 
    & music & $0.344 \pm 0.102$ & $0.705 \pm 0.168$ & \multirow{3}{*}{0.289 / 0.641} \\
    & movie & $0.326 \pm 0.095$ & $0.682 \pm 0.190$ & \\
    & sports & $0.198 \pm 0.100$ & $0.536 \pm 0.259$ & \\
    \cmidrule{1-5}
    \multirow{3}{*}{Llama-3.1-8B} 
    & music & $0.383 \pm 0.089$ & $0.807 \pm 0.111$ & \multirow{3}{*}{0.284 / 0.653} \\
    & movie & $0.315 \pm 0.124$ & $0.686 \pm 0.219$ & \\
    & sports & $0.155 \pm 0.108$ & $0.466 \pm 0.301$ & \\
    \cmidrule{1-5}
    \multirow{3}{*}{OLMo-2-7B} 
    & music & $0.256 \pm 0.119$ & $0.623 \pm 0.228$ & \multirow{3}{*}{0.221 / 0.567} \\
    & movie & $0.241 \pm 0.107$ & $0.603 \pm 0.236$ & \\
    & sports & $0.165 \pm 0.108$ & $0.475 \pm 0.302$ & \\
    \bottomrule
\end{tabular}
\caption{\textbf{Cross-lingual Knowledge Transfer Performance.} 
Each Overall and Transfer score represents the mean $\pm$ standard deviation across all $(L_{\text{train}}, L_{\text{test}})$ language pairs for the given model and domain. The last column shows the average across all domains for each model.}
\label{tab:main_result_avg}
\end{table*}

\begin{figure*}[t]
\centering
\includegraphics[width=\linewidth]{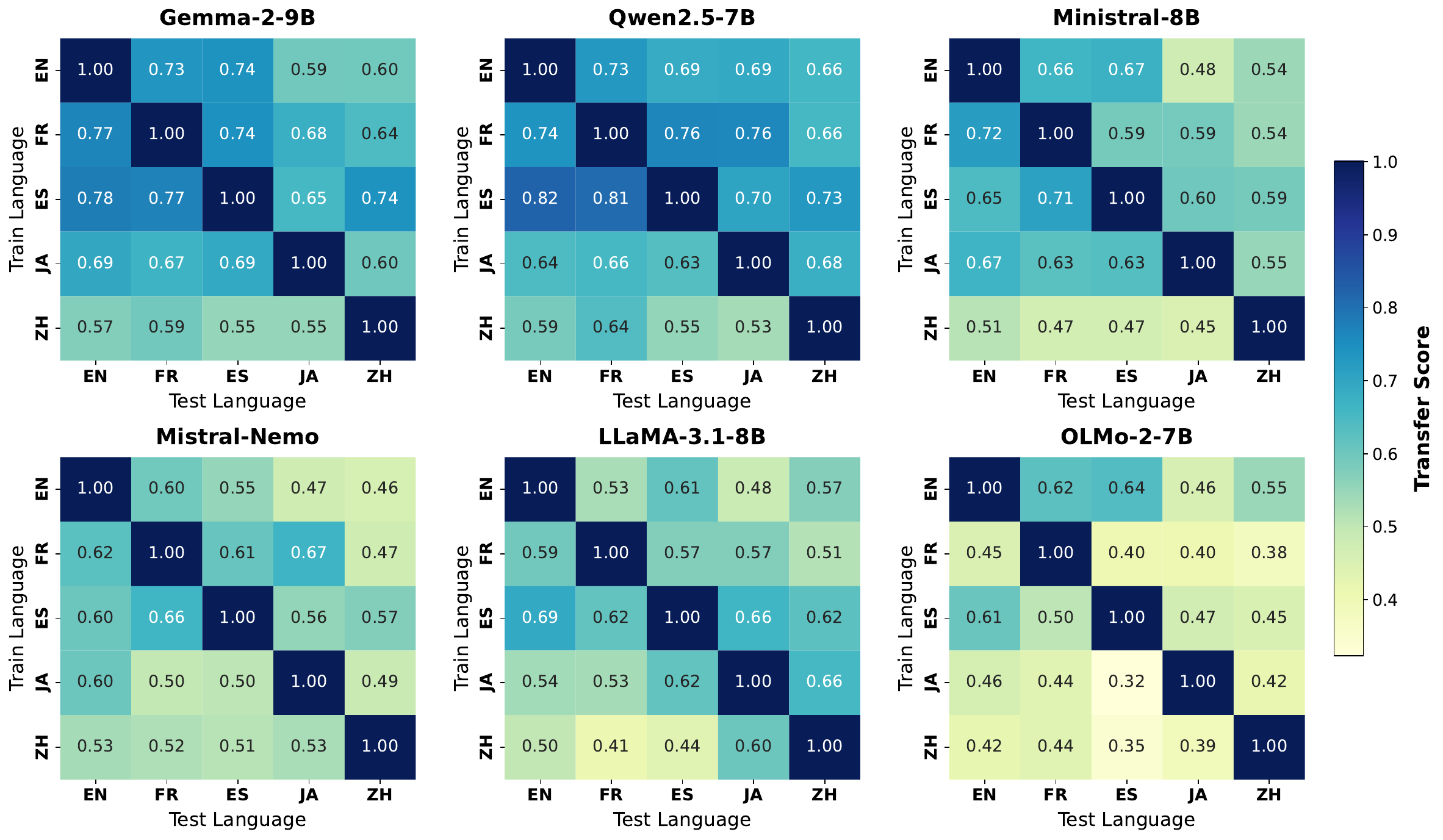}
\caption{\textbf{Language-level Transferability.}
Heatmaps show Transfer Scores for each $(L_{\text{train}}, L_{\text{test}})$ pair across models, sorted by average Overall score. Darker colors indicate stronger transferability.}
\label{fig:transfer_score}
\end{figure*}

\section{Demonstrating LiveCLKTBench: A Case Study}
\label{sec:case_study}
To illustrate the utility of LiveCLKTBench, we conduct a case study demonstrating how it can be applied in practice and its ability to reveal cross-lingual transferability. For simplicity, this demonstration focuses on one of the evaluation scenarios introduced in Section~\ref{sec:protocol}: \emph{Model Comparison}. 

\paragraph{Setup.}  
We evaluate several open-source LLMs using a single knowledge injection strategy: \emph{Continual Pre-Training}~\citep{ke2023continualpretraininglanguagemodels}, where models continue pretraining directly on the target documents using the standard causal language modeling objective (next-token prediction) applied to unmodified text. Although this case study highlights \emph{Model Comparison}, the framework is agnostic to the post-training method and also supports \emph{Strategy Comparison}, enabling evaluation of different knowledge injection techniques for a single model.

\paragraph{Benchmark Configuration.}  
We construct a benchmark covering the period 2025-01-01 to 2025-08-31~\footnote{The most recent model among those evaluated, \textsc{Gemma}, has a pretraining knowledge cutoff of 2024-06.} across five languages: English (en), Japanese (ja), Mandarin (zh), French (fr), and Spanish (es). Each instance consists of training documents in the source language paired with multilingual factual QA test sets, following the generation pipeline described in Section~\ref{sec:construction}. Statistics of the benchmark are shown in Appendix~\ref{app:data_stats}.

\paragraph{Models.}  
We consider instruction-tuned, open-source models in the 7–9B parameter range, including  
\textsc{Gemma}~\footnote{\url{https://huggingface.co/google/gemma-2-9b-it}},  
\textsc{Mistral}~\footnote{\url{https://huggingface.co/mistralai/Ministral-8B-Instruct-2410}}~\footnote{\url{https://huggingface.co/mistralai/Mistral-Nemo-Instruct-2407}},  
\textsc{Qwen}~\footnote{\url{https://huggingface.co/Qwen/Qwen2.5-7B-Instruct}},  
\textsc{Llama}~\footnote{\url{https://huggingface.co/meta-llama/Llama-3.1-8B-Instruct}}, and  
\textsc{OLMo}~\footnote{\url{https://huggingface.co/allenai/OLMo-2-1124-7B-Instruct}}.  
These model families differ in pretraining data, tokenizer design, and alignment strategies, providing a diverse perspective on cross-lingual  transfer.

\paragraph{Training and Inference.}  
All models are post-trained with a lightweight LoRA configuration (rank 16, $\alpha=32$, learning rate $5e^{-4}$, dropout $0.1$, batch size $1$, $5$ epochs). Intermediate checkpoints are validated, and the best-performing one is selected for final testing. At inference, we fix the decoding temperature to $0$ to eliminate randomness and directly reflect the effect of training.

\section{Case Study Results}
\label{sec:experiment}

Building on the setup described in Section~\ref{sec:case_study}, we now present the results of our LiveCLKTBench case study. 
The aim here is not to produce a leaderboard-style evaluation, but to demonstrate the kinds of insights LiveCLKTBench can provide about cross-lingual transfer. 

We first report overall performance across models and domains, then analyze transferability patterns across languages, and examine the effect of model size. 
Together, these results illustrate the diagnostic power of LiveCLKTBench and show that cross-lingual knowledge transfer remains a significant challenge for current LLMs.

\subsection{Overall Performance of Models}
\label{sec:overall}
Table~\ref{tab:main_result_avg} presents the \textit{Overall} and \textit{Transfer} scores for each model across the three domains—music, movies, and sports—as well as their domain averages.
Across models, the average Overall Score ranges from 0.221 (\textsc{OLMo-2-7B}) to 0.414 (\textsc{Gemma-2-9B}), while the Transfer Score ranges from 0.567 (\textsc{OLMo-2-7B}) to 0.747 (\textsc{Qwen2.5-7B}).
\textsc{Gemma-2-9B} achieves the highest average Overall Score, indicating stronger in-language performance, whereas \textsc{Qwen2.5-7B} leads in Transfer Score, suggesting better cross-lingual generalization.
Other models such as \textsc{Ministral-8B}, \textsc{Mistral-Nemo}, and \textsc{LLaMA-3.1-8B} perform moderately across both metrics.

Across domains, most models perform best on \textit{music}, followed closely by \textit{movies}, while \textit{sports} consistently yields the weakest performance.
When comparing models within each domain, their relative rankings remain largely consistent: \textsc{Gemma-2-9B} and \textsc{Qwen2.5-7B} generally outperform others, while \textsc{OLMo-2-7B} trails behind.
This stability of ranking suggests that model-level differences are systematic rather than domain-dependent, indicating comparable generalization patterns across diverse content types.

Overall, these results reveal substantial performance gaps across models, indicating that current LLMs still face challenges in reliably transferring knowledge across languages.

\begin{figure*}[t]
\centering
\includegraphics[width=\linewidth]{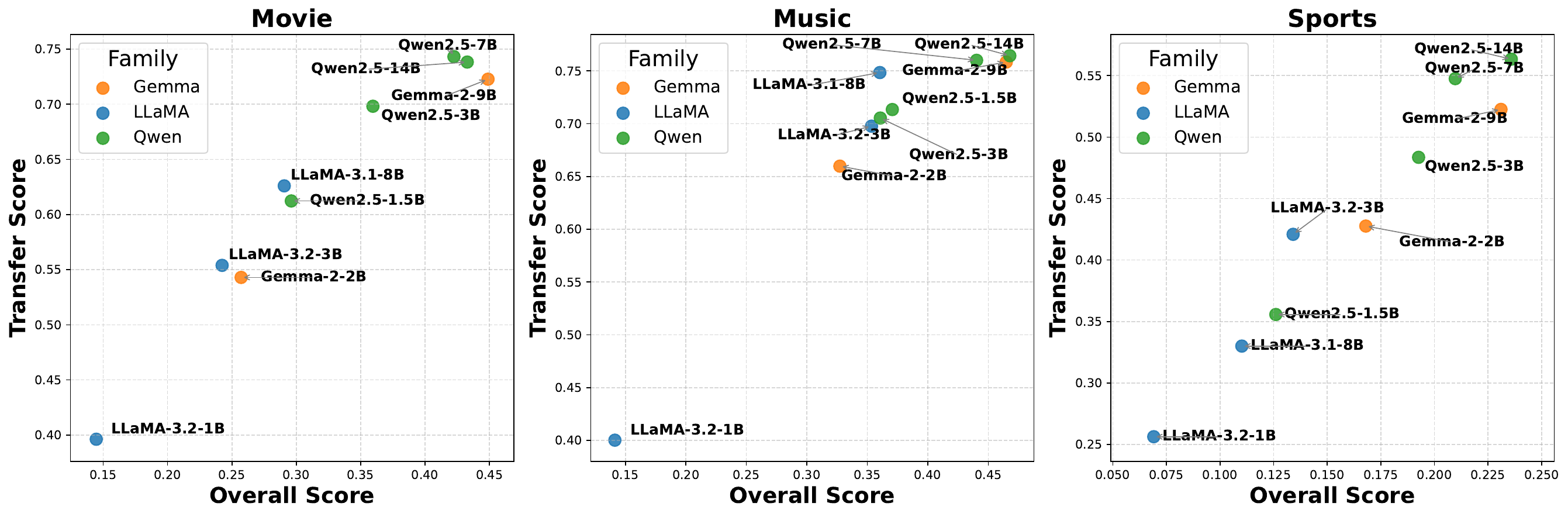}
\caption{\textbf{Effect of Model Size.}
Overall and Transfer Scores across model families of different parameter size, shown separately by domain. }
\label{fig:model_size}
\end{figure*}

\subsection{Variation Across Languages}
\label{sec:lang_trans}

Figure~\ref{fig:transfer_score} presents transfer performance across all source–target language pairs, with models sorted by their Overall score.
Across all models, transferability is consistently weaker for Japanese (ja) and Mandarin (zh) compared to Indo-European languages such as English (en), Spanish (es), and French (fr). However, the degree of degradation varies: stronger models show moderate declines, while weaker ones suffer sharp drops in transfer scores for these distant languages.
This pattern suggests that cross-lingual generalization to typologically different languages remains the key bottleneck, and that model robustness plays an important role in mitigating such gaps.

Another recurring observation is directional asymmetry.
For instance, transferring knowledge from English to Japanese often yields lower performance than from Japanese to English.
Such asymmetries imply that cross-lingual transfer is not bidirectionally balanced and that some languages serve as more effective sources or recipients of knowledge.
Notably, \textsc{Qwen2.5-7B} shows relatively stronger transfer into Mandarin compared to other models, likely reflecting its heavier exposure to Chinese text during pretraining.

Overall, these findings demonstrate that linguistic proximity plays a major role in transfer effectiveness: models achieve higher reliability within language families but face persistent challenges when transferring across typologically distant ones such as Indo-European and East Asian languages.

\subsection{Effect of Model Size}
\label{sec:ablation-size}

Figure~\ref{fig:model_size} illustrates how model size affects transferability across three domains.
Consistent with general scaling trends, larger models consistently outperform smaller ones across all families, confirming that increased capacity generally improves cross-lingual generalization.

However, the improvements are not strictly linear—gains tend to diminish as models grow larger, suggesting a saturation effect at higher scales.
The magnitude of improvement also varies across domains: scaling yields clear benefits in \emph{movies} and \emph{sports}, where larger models achieve noticeably higher transferability, whereas performance on \emph{music} is already high and remains clustered across parameter sizes.

Overall, scaling improves general cross-lingual alignment but with diminishing returns. Beyond a certain capacity, improvements appear constrained more by domain complexity than by model size, suggesting that future gains may require greater efforts toward domain-targeted multilingual adaptation rather than simple scaling.

\section{Conclusion}
\label{sec:conclusion}
In this work, we presented \textbf{LiveCLKTBench}, an automated pipeline for building realistic, contamination-free benchmarks for cross-lingual knowledge transfer. Our approach prevents leakage by selecting independent, time-sensitive knowledge entities, filtering them by temporal occurrence, and discarding any already recognized by pretrained models. The pipeline is fully automated and configurable, enabling frequent updates and flexible customization with minimal human effort. Together, these features make LiveCLKTBench a scalable and sustainable framework for evaluating genuine cross-lingual knowledge transfer in LLMs.

\section*{Limitations}
\label{sec:limit}

While LiveCLKTBench provides a realistic and contamination-free benchmark for evaluating cross-lingual knowledge transfer, several limitations remain that could be addressed in future work:

\textbf{Task Variety.} Currently, the benchmark focuses on multiple-choice QA due to its ease of evaluation and cost efficiency. Extending the framework to include other task types, such as open-ended question answering, would provide a more comprehensive assessment.

\textbf{Domain Coverage.} The benchmark currently spans three domains (movies, music, and sports). Expanding to additional domains and sources would increase knowledge diversity and improve the breadth of evaluation.

\textbf{Language Choices.} In our case study, we evaluate cross-lingual transfer on five languages (English, Spanish, French, Japanese, and Mandarin) based on their coverage across evaluated models, common use in prior multilingual benchmarks, and computational cost considerations. Extending to more languages, especially low-resource ones, is left for future work.

\section*{Ethical considerations}
\label{sec:ethic}
LiveCLKTBench uses publicly available sources such as IMDB/TMDB (movies), YouTube Music (music), and SportsDB (sports) from trusted agencies. While these sources may occasionally contain inaccuracies due to human error or reporting delays, they provide widely recognized and verifiable records of real-world events. All data are derived from post-release or officially published information, ensuring that no private or sensitive material is included. Our benchmark focuses on factual, entity-based knowledge (e.g., movie releases, sports scores), which further minimizes the risk of harmful content.

\section*{Use of Ai Assistants}
\label{sec:ai}
In this work, we leveraged large language models (LLMs) to assist in two ways.
First, LLMs were employed as part of the automatic data generation pipeline for tasks such as question formulation, translation, and quality verification.
Second, an AI assistant (OpenAI GPT-5) was used for minor writing support, including grammar correction and improving manuscript clarity.
All AI-assisted content was carefully reviewed and verified by the authors to ensure factual accuracy and alignment with the authors’ original intent.

\section*{Acknowledgment}
\label{sec:ack}
This material is based upon work supported by National Science and Technology Council, ROC under grant number 114-2221-E-002-134-MY3 and 113-2628-E-001-003-MY4, NTU AI Center of Research Excellence within Taiwan Centers of Excellence in Artificial Intelligence, and by National Taiwan University and Academia Sinica Innovative Joint Program, under grant AS-NTU-114-06.

\bibliography{latex/custom}

\appendix

\onecolumn

\section{Data Statistics}
\label{app:data_stats}
In LiveCLKTBench, the training set consists of raw text documents that contain information about each knowledge entity (e.g., a movie). Each training document is associated with multiple QA pairs that probe different factual aspects of the same entity (e.g., cast, storyline, release details).
Since we evaluate cross-lingual transfer by post-training the model on the source-language documents, we split the corresponding QA pairs into validation (20\%) and test subsets (80\%). The validation subset is used to select the best checkpoint and to monitor learning progress during post-training. The test subset is held out and used only for final evaluation. Importantly, both validation and test questions are still grounded in the same knowledge document used for training and do not introduce any new information.

\begin{table}[htbp]
\centering
\begin{tabular}{lcccc}
\toprule
\textbf{Domain} & \textbf{\#Entities} & \textbf{\#Fact QAs (pre-split)} & \textbf{\#Validation} & \textbf{\#Test} \\
\midrule
Movie  & 30 & 175 & 175 & 700 \\
Music  & 30 & 125 & 125 & 500 \\
Sports & 20 & 95  & 90  & 380 \\
\bottomrule
\end{tabular}
\caption{\textbf{Benchmark Statistics.}
Number of entities, generated factual QA pairs before translation and splitting, and counts of the final translated validation and test instances.}

\label{tab:data_stats}
\end{table}

\section{Quality of LLM QA Generation}
\label{app:data_verify}
To assess the reliability of our benchmark data, we recruited three annotators (two undergraduate students and one graduate student) to manually verify the quality of sampled question–answer pairs.
Specifically, we randomly sampled 20\% of entities from each domain and evaluated all QA pairs associated with them.
Each question was independently annotated by three annotators based on the following criteria, and was considered a failure if it violated any of them.

\begin{itemize}
    \item F-1: The question is unrelated to the source document.
    \item F-2: The question is relevant but overly general and not specific to the target entity.
    \item F-3: The answer cannot be verified from the source document.
    \item F-4: The answer is verifiable from the source document but is incorrect.
\end{itemize}

For annotation, we required full agreement among annotators. Questions with disagreement were discarded from the analysis.
On the remaining items (98.8\%), the validated QA accuracy (Pass Rate) is 95.2\%.
The invalid items include 3.6\% cases of F-3 (from the movie domain) and 1.2\% cases of F-4 (from the sports domain).
These results suggest that our automated pipeline produces reasonably high-quality, fact-grounded QA pairs.

\section{Quality of LLM Translation}
\label{app:translation_quality}
LiveCLKTBench aims to support continuous, contamination-controlled updates, making automatic translation essential for scalability. Consistent with prior multilingual benchmarks~\cite{lai2023okapiinstructiontunedlargelanguage, goldman2025eclektic, asai2023buffetbenchmarkinglargelanguage, singh2025globalmmluunderstandingaddressing, zhang2023m3exammultilingualmultimodalmultilevel}, we adopt LLM-based translation. Our pipeline utilizes GPT-4o-mini~\cite{openai2024gpt4o}, a strong multilingual model comparable to those used in prior work, ensuring that translation quality is aligned with existing benchmarks.

To further validate translation quality, we conduct human evaluation on a randomly sampled 10\% subset of QA pairs and training documents across four target languages: English, Japanese, French, and Mandarin. These languages are selected based on the availability of native or highly proficient annotators. Each sample is evaluated by annotators who are either native speakers or have advanced proficiency in the target language, using two 1--5 scales (higher is better):

\begin{itemize}
    \item \textbf{Adequacy (semantic correctness):} How accurately the original meaning is preserved in the translation.
    \item \textbf{Fluency (linguistic naturalness):} How natural, grammatical, and fluent the translation is in the target language, independent of semantic accuracy.
\end{itemize}

\begin{table}[h]
\centering
\small
\begin{tabular}{lcccc}
\toprule
\textbf{Language Pair} & \textbf{Adequacy (Train Doc)} & \textbf{Adequacy (Test QA)} & \textbf{Fluency (Train Doc)} & \textbf{Fluency (Test QA)} \\
\midrule
EN $\rightarrow$ JA & 5.00 & 4.67 & 3.88 & 3.86 \\
EN $\rightarrow$ FR & 4.88 & 4.92 & 4.50 & 4.72 \\
EN $\rightarrow$ ZH & 4.75 & 5.00 & 4.25 & 4.67 \\
\bottomrule
\end{tabular}
\caption{Human evaluation results for translation quality across language pairs.}
\label{tab:translate}
\end{table}

Table~\ref{tab:translate} shows consistently high adequacy scores across all language pairs, indicating that the translated content reliably preserves the original meaning. Fluency scores are comparatively lower, suggesting that there is room for improvement in linguistic naturalness. Nevertheless, the strong adequacy scores support the validity of the multilingual QA pairs, since the factual content required for evaluation is preserved even when surface-level naturalness varies.
To further improve fluency in future iterations, we plan to adopt stricter translation constraints (e.g., discouraging overly literal phrasing), incorporate back-translation techniques, and explore lightweight human-in-the-loop refinement for low-fluency cases.

\section{QA Examples}
\label{app:qa_example}
\begin{examplebox}{Music}
\texttt{In the music video 'Alex Warren - Ordinary (Official Video)', what recurring theme is mentioned in the lyrics?}  \\
(A.) Love and longing 
(B.) Self-empowerment and resilience 
(C.) Escaping from fame and pressure 
(D.) Chasing dreams and freedom
\end{examplebox}
\begin{examplebox}{Movie}
\texttt{In the movie 'KPop Demon Hunters', who are the main characters that use their secret identities to fight supernatural threats?}  \\
(A.) Zoey, Mira, Ahn  
(B.) Arden, May, Ji-young  
(C.) Rumi, Mira, Zoey 
(D.) Ahn, Yunjin, Rumi  
\end{examplebox}

\section{Entity Document Examples}
\label{app:doc_metadata}

\begin{tcolorbox}[promptbox, title={Movie Document Template}]
    \texttt{- Movie Title: \{title\}} \\ 
    \texttt{- Movie Cast: \{casts\}} \\ 
    \texttt{- Movie Summary: \{summary\}} \\ 
    \texttt{- Movie Synopsis: \{synopsis\}} 
\end{tcolorbox}

\begin{tcolorbox}[promptbox, title={Music Document Template}]
    \texttt{- Music Video Title: \{title\}} \\ 
    \texttt{- Music Release Date: \{date\}} \\ 
    \texttt{- Music Video Description: \{description\}} 
\end{tcolorbox}


\begin{tcolorbox}[promptbox, title={Sports Document Template}]
\texttt{Sports: \{sports\}} \\ 
\texttt{League: \{league\}} \\[0.5em]

\texttt{Match: \{home\_team\} vs \{away\_team\}} \\ 
\texttt{Date: \{date\}} \\ 
\texttt{Score: \{home\_score\} - \{away\_score\}} \\ 
\texttt{Venue: \{venue\}} \\[0.5em]

\texttt{Innings Breakdown:} \\ 
\texttt{\{home\_team\}: \{home\_innings\} → Hits: \{home\_hits\}, Errors: \{home\_errors\}} \\ 
\texttt{\{away\_team\}: \{away\_innings\} → Hits: \{away\_hits\}, Errors: \{away\_errors\}} 
\end{tcolorbox}


\section{Pipeline Prompt Examples}
\label{app:prompts}

\begin{tcolorbox}[promptbox, title={QA GENERATION PROMPT}]

\texttt{You are generating high-quality multiple-choice QA pairs in \{lang\}, strictly grounded in the given movie information. \\
\\
You will be provided with: \\
- Movie Title \\
- Movie Casts \\
- Movie Summary \\
- Movie Synopsis \\
\\
Task: \\
- Generate natural, audience-friendly questions that viewers might realistically ask. \\
- All questions must be written fully in \{lang\}, including the leading phrase (“In the movie: '<title>', …”). \\
- Each QA pair must be based ONLY on facts explicitly present in the input. Do not add, assume, or hallucinate. \\
- Use diverse aspects (casts, summary, synopsis content). \\
\\
Each QA pair must include: \\
- Question in \{lang\}, beginning with “In the movie: '<title>', …” (do not translate title) \\
- Options: \\
    - Provide four options labeled A, B, C, D. \\
    - Exactly one option is correct. \\
    - Place the correct option randomly among A–D (do not always use the same position). \\
    - Distractors must be plausible but wrong (no random, absurd, or unrelated answers). \\
- Correct Option: Output the letter (A, B, C, or D) of the correct answer. \\
\\
-------------------------- \\
Inputs: \\
\{meta\_data\} \\
\\
-------------------------- \\
Output Format (JSON): \\
\{ \\
    "QA": [ \\
        \{ \\
            "question": "<string in \{lang\}>", \\
            "options": \{ \\
                "A": "<option A in \{lang\}>", \\
                "B": "<option B in \{lang\}>", \\
                "C": "<option C in \{lang\}>", \\
                "D": "<option D in \{lang\}>" \\
            \}, \\
            "correct\_option": "<A | B | C | D>" \\
        \}, \\
        ... \\
    ] \\
\} \\
\\
-------------------------- \\
Guidelines: \\
- Write everything (questions and options) only in \{lang\}. \\
- Keep all proper names (people, places, entities) unchanged. \\
- Ensure every correct answer can be directly verified in the input metadata. \\
- Distractors must be reasonable, related, and plausible. \\
}
\end{tcolorbox}

\begin{tcolorbox}[promptbox, title={QA VERIFIER PROMPT}]
\texttt{You are verifying if QA pairs are grounded in the provided metadata. \\
\\
Check: \\
- Is the correct option explicitly supported by the metadata? \\
- If yes, return SUPPORTED and the supporting sentence(s). \\
- If not, return UNSUPPORTED. \\
\\
Output Format: \\
\{ \\
\ \ "Decision": "<SUPPORTED or UNSUPPORTED>", \\
\ \ "SourceSentence": "<sentence(s) from metadata or empty>" \\
\}}
\end{tcolorbox}

\begin{tcolorbox}[promptbox, title={QA TRANSLATION PROMPT}]
\texttt{Translate the QA JSON into target language \{lang\}. \\
\\
Rules: \\
- Translate only values of "question" and "options". \\
- Do NOT translate keys ("QA", "question", "options", "A"..."D", "correct\_option"). \\
- Keep "correct\_option" unchanged. \\
- Preserve JSON structure.}
\end{tcolorbox}

\begin{tcolorbox}[promptbox, title={DOCUMENT TRANSLATION PROMPT}]
\texttt{Translate the following movie document into \{lang\}. \\
- Do NOT translate the field names (e.g., "Movie Cast", "Movie Summary", "Movie Synopsis"). \\
- Translate only the values after the colon. \\
- If the text is already in \{lang\}, return it unchanged. \\
- Return only the JSON object, no extra text. \\
\\
Document: \\
- Movie Cast: \{casts\} \\
- Movie Summary: \{summary\} \\
- Movie Synopsis: \{synopsis\} \\
\\
Output Format: \\
\{ \\
\ \ "translation": \{ \\
\ \ \ \ "Cast": "<translated cast>", \\
\ \ \ \ "Summary": "<translated summary>", \\
\ \ \ \ "Synopsis": "<translated synopsis>" \\
\ \ \} \\
\}}
\end{tcolorbox}

\end{document}